\documentclass[lettersize,journal]{IEEEtran}

\usepackage{amsmath,amsfonts}
\usepackage{algorithm}
\usepackage{array}
\usepackage{textcomp}
\usepackage{stfloats}
\usepackage{url}
\usepackage{verbatim}
\usepackage{graphicx}

\usepackage{multirow}
\usepackage{makecell}
\usepackage{adjustbox}
\usepackage{booktabs}
\usepackage[numbers]{natbib}
\usepackage{amssymb}

\usepackage{booktabs}
\usepackage[caption=false,font=normalsize,labelfont=sf,textfont=sf]{subfig}
\usepackage{algpseudocode}
\usepackage{bbding}
\usepackage{color}
\usepackage[switch]{lineno}
\graphicspath{ {./imgs/} }

\usepackage{xspace}
\usepackage[colorlinks]{hyperref}

\begin{document}
\title{Few-shot Oriented Object Detection with Memorable Contrastive Learning in Remote Sensing Images}

\author{
\IEEEauthorblockN{
Jiawei Zhou\textsuperscript{1,4}, 
Wuzhou Li\textsuperscript{2}, 
Yi Cao\textsuperscript{1}, 
Hongtao Cai\textsuperscript{1,4}, 
Xiang Li\textsuperscript{3*}
}

\IEEEauthorblockA{\textsuperscript{1}Electrical Infomation School, Wuhan University, Wuhan, 430072 China}

\IEEEauthorblockA{\textsuperscript{2}School of Remote Sensing and Information Engineering, Wuhan University, Wuhan, 430072 China}

\IEEEauthorblockA{\textsuperscript{3}King Abdullah University of Science and Technology, Thuwal, 23955, Saudi Arabia}

\IEEEauthorblockA{\textsuperscript{4}Hubei Luojia Laboratory, Wuhan, 430072 China}

\thanks{Corresponding author: Xiang Li (email: xiangli92@ieee.org). The Project is supported by the Special Fund of Hubei Luojia Laboratory (No.220100013)
}
}

\IEEEtitleabstractindextext{\begin{abstract}
Few-shot object detection (FSOD) has garnered significant research attention in the field of remote sensing due to its ability to reduce the dependency on large amounts of annotated data. However, two challenges persist in this area: (1) axis-aligned proposals, which can result in misalignment for arbitrarily oriented objects, and (2) the scarcity of annotated data still limits the performance for unseen object categories. To address these issues, we propose a novel FSOD method for remote sensing images called Few-shot Oriented object detection with Memorable Contrastive learning (FOMC). Specifically, we employ oriented bounding boxes instead of traditional horizontal bounding boxes to learn a better feature representation for arbitrary-oriented aerial objects, leading to enhanced detection performance. To the best of our knowledge, we are the first to address oriented object detection in the few-shot setting for remote sensing images. To address the challenging issue of object misclassification, we introduce a supervised contrastive learning module with a dynamically updated memory bank. This module enables the use of large batches of negative samples and enhances the model's capability to learn discriminative features for unseen classes. We conduct comprehensive experiments on the DOTA and HRSC2016 datasets, and our model achieves state-of-the-art performance on the few-shot oriented object detection task. Code and pretrained models will be released.
\end{abstract}

}

\maketitle

\IEEEdisplaynontitleabstractindextext

\IEEEpeerreviewmaketitle

\def \stanet {${{\rm{S}}^{\rm{2}}}{\rm{A}}$-Net\xspace}

\section{Introduction}

\IEEEPARstart{O}{bject} detection for remote sensing images is a fundamental task,  widely applied in many real-world applications \cite{cheng2016survey, ok2015circular, yu2015vehicle, sirmacek2010probabilistic,hu2019sample}. In recent years, advancements in deep learning-based methodologies have significantly enhanced the domain of object detection, primarily due to their intricate architectures and robust feature extraction capabilities \cite{girshick2014rich}. Despite the breakthrough, existing models predominantly rely upon substantial quantities of annotated training data to achieve satisfactory performance, a stipulation that introduces noteworthy impediments \cite{liu2017high, xia2018dota}. The acquisition and annotation of requisite remote sensing data entail considerable expenditures of both financial and human resources. This limitation hinders the broader adoption of these techniques in practical applications \cite{sun2021research}. Notably, the remarkable human capacity for swift recognition of novel objects with only a handful of exemplars has catalyzed the investigation into the domain of few-shot object detection (FSOD) within the context of remote sensing imagery~\cite{li2021few,wolf2021double,cheng2021prototype,zhou2022few}. This emerging field focuses on the detection of previously unseen objects using only a handful of annotated samples for training. The burgeoning interest in this area underscores its potential significance. However, the field is currently contending with two primary challenges:

\begin{figure}
  \centering
  \includegraphics[width=9.0cm]{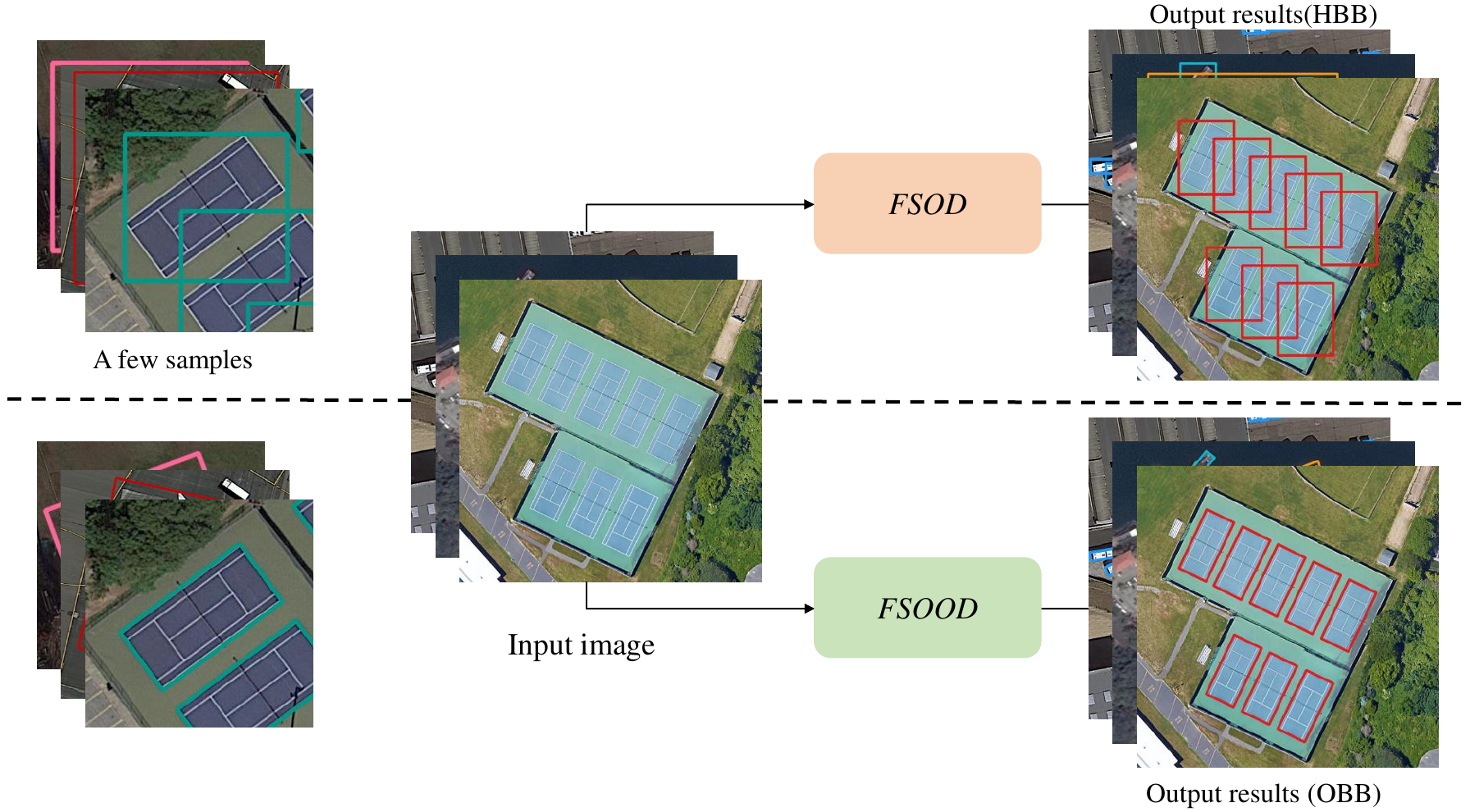}
  \caption{Comparison between the few-shot object detection (FSOD) task predicts HBBs (top) and the newly proposed few-shot oriented object detection (FSOOD) task that outputs OBBs (bottom). HBBs often cover background areas and adjacent objects, particularly for dense or large-scale objects. Conversely, the OBBs provide a more accurate representation of objects with tighter bounding boxes.
  }\label{fig1}
\end{figure}
\begin{enumerate}
\item \textbf{Arbitrarily oriented objects.} Unlike natural scene images, remote sensing images are generally taken with a bird's-eye view. Therefore, foreground objects typically appear in arbitrary orientations. Existing FSOD methods for remote sensing images have been designed to accommodate axis-aligned bounding box detection, which may inadvertently cover adjacent instances and complex background areas. For instance, as shown in Fig. \ref{fig1}, objects like vehicles in parking lots or ships in harbors can be densely packed. The traditional axis-aligned horizontal bounding boxes (HBBs) typically encompass multiple objects, causing ambiguity for the subsequent classification and location task, especially in situations of limited training data. Moreover, objects like bridges vary widely in scale and aspect ratio. This variability further increases the challenge of accurately localizing them, especially with limited training data.

\item \textbf{Misclassification of novel instances.} In few-shot detection, misclassification of novel object categories can be the main source of errors rather than localization \cite{sun2021fsce}. Learning a well-separated decision boundary for novel categories remains a challenging issue due to the lack of training data. 
\end{enumerate}

To address the identified challenges, this work introduces a pioneering approach named Few-shot Oriented object detection with Memorable Contrastive learning (FOMC). This method tackles the first challenge by employing oriented bounding boxes (OBBs) instead of horizontal bounding boxes (HBBs) in few-shot detection tasks. Refer to Fig. \ref{fig1} for a visual comparison. OBBs are particularly effective for aerial object detection as they offer improved object alignment and significantly reduce background interference while preserving the orientation details of the objects. To this end, a novel task, termed Few-Shot Oriented Object Detection (FSODD) is proposed, which focuses on accurately predicting OBBs for previously unseen object categories using a limited number of annotated examples.

To address the second challenge, we introduce a Memorable Contrastive Learning (MCL) module designed to enhance the model's ability to extract distinctive feature representations. This module leverages insights from recent developments in contrastive learning. The fundamental principle of contrastive learning is to ensure that positive (similar) pairs are closely mapped in the embedding space, while negative (dissimilar) pairs are positioned further apart. The selection and utilization of negative samples are crucial in contrastive learning methodologies. As evidenced by existing research \cite{He2020Moco,khosla2020supervised}, incorporating a greater variety of negative samples improves the distinction between signal (positive) and noise (negative) data. Inspired by this, we adapt the self-supervised framework MoCo \cite{He2020Moco} to a fully-supervised context for better class-separable feature learning in the newly proposed FSOOD task. Furthermore, to solve label confusion arising from unselected shots (objects) in images with partially selected shots, we introduce a novel shot masking technique to eliminate unselected shots for better supervision in few-shot detection task.

Our main contributions can be summarized as follows:
\begin{enumerate}
\item We introduce a new task termed few-shot oriented object detection, which focuses on predicting oriented bounding boxes for unseen object categories with only a few annotated samples. A baseline method is developed to solve the task.
\item We propose an MCL module, a contrastive learning-based approach with a dynamically updated memory bank. This mechanism guides the model to learn more discriminative features with larger intra-class compactness and inter-class differences by covering a richer set of negative samples across different mini-batches.
\item A shot masking technique is developed to mitigate the label confusion arising from unselected shots (objects) in images with partially selected shots, significantly enhancing our model's performance.
\item  We conduct experiments on two widely-used benchmarks, including the DOTA~\cite{xia2018dota} and HRSC2016~\cite{liu2016ship} datasets. Results show that FOMC achieves state-of-the-art performance in few-shot oriented object detection on remote sensing images.
\end{enumerate}

\section{RELATED WORKS}

\subsection{Object detection in remote sensing images}

For common object detection, deep learning-based methods can be roughly categorized as two-stage detectors and one-stage detectors. Two-stage detectors generate region proposals that potentially contain objects and then refine the location and make the final prediction based on region proposals.
Representative two-stage detectors include R-CNN \cite{girshick2014rich}, SPPNet \cite{he2015spatial}, Fast R-CNN \cite{girshick2015fast}, Faster R-CNN \cite{ren2015faster} and Feature Pyramid Network (FPN) \cite{lin2017feature}. In contrast, one-stage detectors complete the object detection prediction in one step without pre-defined anchors. YOLO \cite{redmon2016you}, SSD \cite{liu2016ssd}, RetinaNet \cite{lin2017focal}, Cornernet \cite{law2018cornernet}, CenterNet \cite{duan2019centernet} and transformer-based DETR \cite{carion2020end} are representative one-stage detectors.
In the remote sensing field, object detection is a fundamental task that detects instances of a certain class (ships, vehicles, or airplanes) in aerial images. Driven by the great success in the common object detection field, researchers have adopted deep learning-based methods to address the detection problem in remote sensing images. Li et al. \cite{li2017rotation} propose a rotated-insensitive RPN by adding multi-angle anchors to generate translation-invariant region proposals. Zhuang et al.\cite{zhuang2019single} propose a single-shot detector with a multi-scale feature fusion module that can utilize both the semantic information from high-level features and fine details from low-level features. Cheng et al. \cite{cheng2020cross} extend the Faster R-CNN by designing a cross-scale feature fusion module to extract powerful multi-level features for remote sensing images.

\subsection{Oriented object detection in remote sensing images}
Numerous studies have been conducted to investigate the challenge of oriented object detection within the realm of remote sensing imagery. Liu et al. \cite{liu2017rotated} propose the rotated region-based CNN (RR-CNN) to accurately locate rotated ships. Li et al. \cite{li2018multiscale} also focus on ship detection, employing a five-parameter method to describe oriented bounding boxes for ships in complex conditions. Ding et al. \cite{ding2018learning} introduce a rotated Region of Interest (RoI) learner, which transforms horizontal RoIs into oriented RoIs for the detection of oriented objects. Yang et al. \cite{yang2019scrdet} introduce SCRDet to detect small, cluttered, and rotated objects. Specifically, they devised a sampling fusion network to combine multi-layer features for detecting small objects. Additionally, they implement an attention-based network to fuse multi-layer features, and for rotated objects, they incorporate the Intersection over Union (IoU) constant factor into the smooth L1 loss. Qing et al. \cite{qing2021improved} introduce a YOLO-based network with an improved FPN and a path-aggregation module. This work aims to detect free-angle aerial targets with better inference efficiency. Hou et al. \cite{hou2021self} propose a self-adaptive aspect ratio anchor to explore variations in aspect ratios and employ an oriented box decoder for effective encoding of orientation information in oriented object detection. Han et al. \cite{han2021align} introduce the Single-Shot Alignment Network (\stanet) tailored for oriented object detection. Comprising two modules—a Feature Alignment Module (FAM) and an Oriented Detection Module (ODM)—\stanet serves as our chosen baseline method in this study. Its framework allows seamless integration of our proposed methods in a plug-and-play manner, striking a favorable balance between training accuracy and inference speed.

\subsection{Few-shot object detection}
Current FSOD methods can be grouped into two categories: meta-learning-based \cite{li2021few, wang2019meta}, and transfer learning-based \cite{wang2020frustratingly, fan2021generalized, cheng2021prototype}. Meta-learning-based approaches acquire the knowledge about ``learning to learn'' by designing and solving a series of FSOD tasks, enabling a model to adapt and generalize to novel data more efficiently. This process typically involves a complex data-organization paradigm and training process, making it challenging to apply in practical scenarios. On the other hand, transfer learning-based methods aim to leverage knowledge learned from sufficient data and transfer it by fine-tuning the model with a small amount of novel data. Notably, TFA \cite{wang2020frustratingly} adopts a two-stage ``pretrain-finetune'' paradigm: pretrain the model on sufficient base data and then only fine-tune the last few layers with novel data. Despite its simple framework and training strategy, TFA outperforms many meta-learning approaches~\cite{wang2020frustratingly}. In this study, we adopt the two-stage fine-tuning training strategy to achieve simple and effective FSOD for remote sensing images.

\subsection{Few-shot object detection in remote sensing images}
FSODM \cite{li2021few} is one of the pioneering works exploring FSOD in remote sensing. In this work, they develop a multi-scale object detector with a feature reweighting module that assigns different weights to features from support images. Wolf et al. \cite{wolf2021double} introduce a double-head predictor to decouple the prediction for base classes and novel classes, preventing performance degradation. Cheng et al.\cite{cheng2021prototype} propose a prototype-CNN to convert support images to class-specific prototypes, and fuse the information to generate better proposals, further enhancing detection performance. Xiao et al. \cite{xiao2021few} present a self-adaptive attention network to fully leverage the instance-level relations for better detection results. Zhou et al.\cite{zhou2022few} develop an FSOD model for scale-variant remote sensing images by incorporating context-aware pixel aggregation and context-aware feature aggregation modules to extract semantic information from different scales. Zhao et al.\cite{zhao2021few} utilize the involution operator in the backbone network to enhance the classification performance and then designed a path-aggregation module for multi-scale detection. Li et al.\cite{li2023few} leverage metric learning and develop a multi-similarity module to learn distinctive feature representations, achieving impressive FSOD performance on aerial imagery. However, these methods generally overlook the detection of rotated objects with OBBs, a crucial consideration in remote sensing images. Our work aims to address the few-shot oriented object detection challenge for objects in aerial imagery.

\subsection{Contrastive learning}
The fundamental concept underlying contrastive learning is elegantly simple: the objective is to learn robust representations by minimizing the distances between positive samples while simultaneously maximizing distances between negative ones in the embedding space. This optimization process enhances the quality of learned features. Prior self-supervised contrastive learning methods take differently augmented views of the same data as positive samples, and randomly picked objects from the mini-batch as negative samples. For instance, SimCLR \cite{chen2020Simclr} introduces a straightforward framework that computes pairwise similarities among images within the mini-batch. On the other hand, MoCo \cite{He2020Moco} devises a momentum-updated dictionary with a queue, enabling the utilization of considerably large batches of negative samples. The increasing number of negative samples improves the discrimination ability between positive and negative samples, as demonstrated in several papers \cite{He2020Moco,chen2020moco2,khosla2020supervised}. 

Supervised contrastive learning methods can leverage label information to achieve superior performance. These methods regard positive samples as those with the same label, and samples from different classes naturally serve as negatives. However, most prior research focuses on image classification \cite{khosla2020supervised} and identification \cite{sun2014deep}, and merely a few works adopt it for the FSOD task. FSCE \cite{sun2021fsce} proposes a contrastive proposal encoding branch to learn discriminative feature embeddings and enhance the detection performance for natural scene images. Huang et al. \cite{huang2022few} propose an FSOD model with dual-contrastive learning which consists of a supervised contrastive branch and a prototypical contrastive learning module to learn good representations to boost the performance for classification and then reserve the learned information to maintain the performance for the base classes. In the few-shot setting, positive samples are scarce for novel instances, making the full utilization of negative samples crucial for the model to learn discriminative features. Motivated by this insight, our study draws inspiration from the self-supervised MoCo method, adapting it to the FSOD task to address this specific challenge.

\label{sec:3.1}

\begin{figure*}
	\centering
	\subfloat[Base training stage]{\includegraphics[scale=0.7]{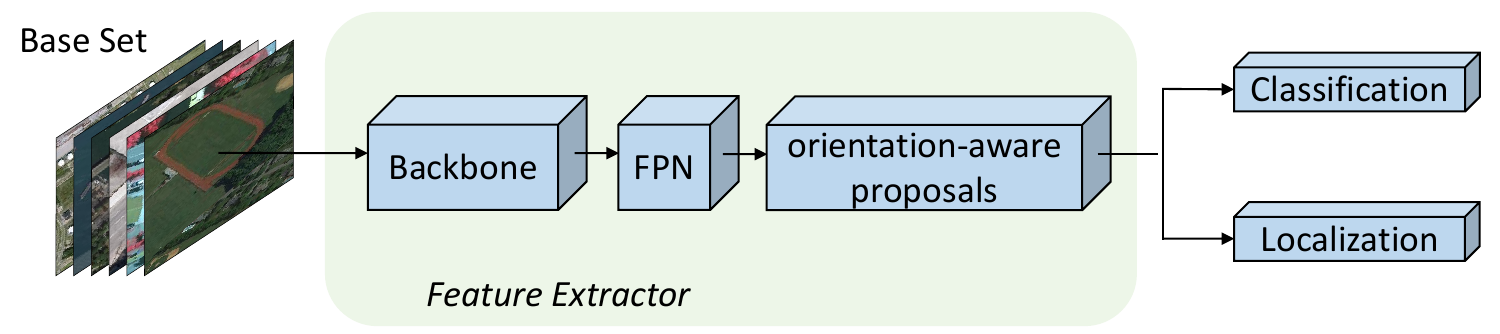}}
        \newline
	\subfloat[Fine-tuning stage]{\includegraphics[scale=0.7]{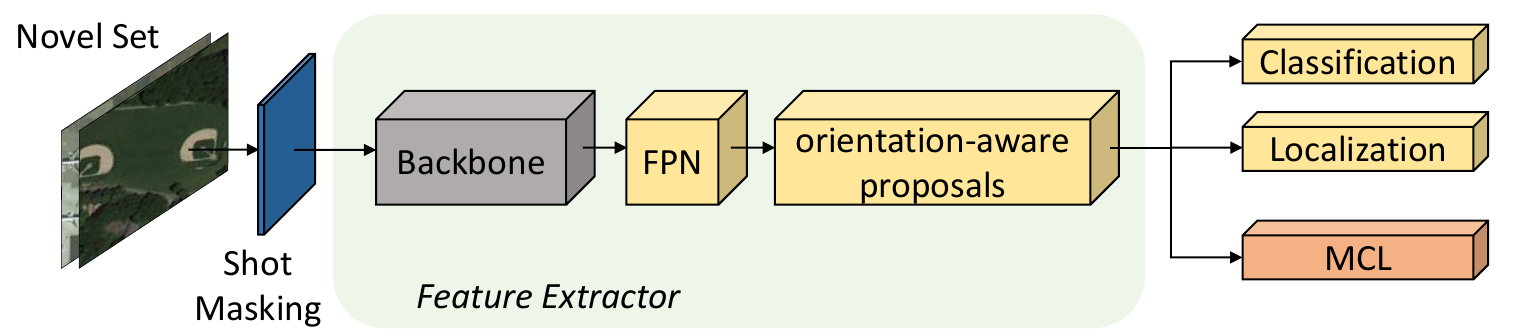}}
	\caption{The overall architecture of our proposed FOMC model. (a) In the base training phase, all network parameters of the model are trained using abundant data from the base categories. (b) In the fine-tuning stage, the parameters of the ResNet backbone network are frozen, while the other modules are trained at a lower learning rate using a few samples from novel categories. An MCL module is designed to store encoded proposal features for contrastive learning and encourage the model to learn class distinctive features.}
    \label{fig2}
\end{figure*}

\section{Methodology}
We introduce our FOMC method in the following. Section \ref{sec_problem} details the newly introduced few-shot oriented object detection task. Section \ref{sec_overview} provides an overview of our proposed method. The MCL module and the shot masking strategy are discussed in Section \ref{sec_MCL} and Section \ref{sec_masking}, respectively. Finally, the training strategy is described in Section \ref{sec_training}.

\subsection{Few-shot Oriented Object Detection (FSOOD)}\label{sec_problem}
We first introduce the settings for the few-shot oriented object detection problem. Our goal is to obtain a detection model that learns from abundant data from base categories and a few annotated data from novel categories. This model should be capable of effectively detecting both base objects and novel objects and should generate OBBs rather than the traditional HBBs for accurate object localization. 
To this end, the categories of the datasets are divided into base categories $C_\text{base}$ and novel categories $C_\text{novel}$. Consequently, the dataset can be segmented into $D_\text{base}$ and $D_\text{novel}$ sub-datasets. $D_\text{base}$ sub-dataset contains all data belonging to base categories $C_\text{base}$, while $D_\text{novel}$ sub-dataset is a small dataset containing equally sampled shots from all categories, including $C_\text{base}$ and $C_\text{novel}$. 

\subsection{Method Overview}\label{sec_overview}
Our FOMC employs a two-stage training paradigm, as shown in Fig. \ref{fig2}. In the base training stage, the base model is trained with an abundant amount of data from base categories. In this work, we use \stanet as the base model, based on two merits: (1) It is a one-stage object detection framework with a better trade-off between high efficiency and accuracy; (2) It generates orientation-sensitive features for regression and orientation-invariant features for classification, achieving good results in the classification and localization of rotated objects.

Given input aerial images, a backbone network along with a feature pyramid network (FPN) is used to generate feature maps at multiple scales. To encourage orientation-aware feature alignment, we leverage a feature alignment module (FAM) following ~\cite{han2021align}. In FAM, an anchor refinement network (ARN) is used to generate high-quality rotated anchors, followed by an alignment convolution layer (ACL) to extract orientation-aligned features. In the detection module, we follow \cite{han2021align} to use ARF \cite{zhou2017oriented} to generate orientation-sensitive features, followed by a pooling operation to obtain orientation-invariant features. The orientation-sensitive features are fed into the regression head to predict orientated bounding box coordinates; while orientation-invariant features are fed into the classification head to predict object classification scores.

In the fine-tuning stage, the base model is further finetuned using a few annotated samples from both base and novel categories. Specifically, a small balanced dataset is created by randomly sampling $K$ ({$K$ = 3, 5, 10, 20}) shots of instances from both $C_\text{base}$ and $C_\text{novel}$ categories. In this stage, the backbone network is frozen to prevent overfitting, while the other modules, including FPN, feature alignment module, object detection module, as well as newly designed MCL module are trained simultaneously to learn object detection on both base and novel categories.

\subsection{Memorable Contrastive Learning module}\label{sec_MCL}
Recent studies like~\cite{sun2021fsce} introduce a contrastive proposal encoding loss that facilitates instance-level intra-class compactness and inter-class separability. However, this approach computes the contrastive loss exclusively on proposals within each iteration. The prevailing wisdom in the field of contrastive learning, as suggested by works like \cite{He2020Moco,chen2020moco2,chen2020simple}, posits that larger batch sizes are essential for effective feature learning. To address this limitation, our work draws inspiration from the well-established contrastive learning framework MoCo \cite{He2020Moco}. We innovate by designing a proposal memory bank that allows for the computation of contrastive losses across multiple iterations. This strategy significantly increases the variety of samples available for contrastive learning, thereby enhancing the overall learning efficacy.

\begin{figure}
  \centering
  \includegraphics[width=0.48\textwidth]{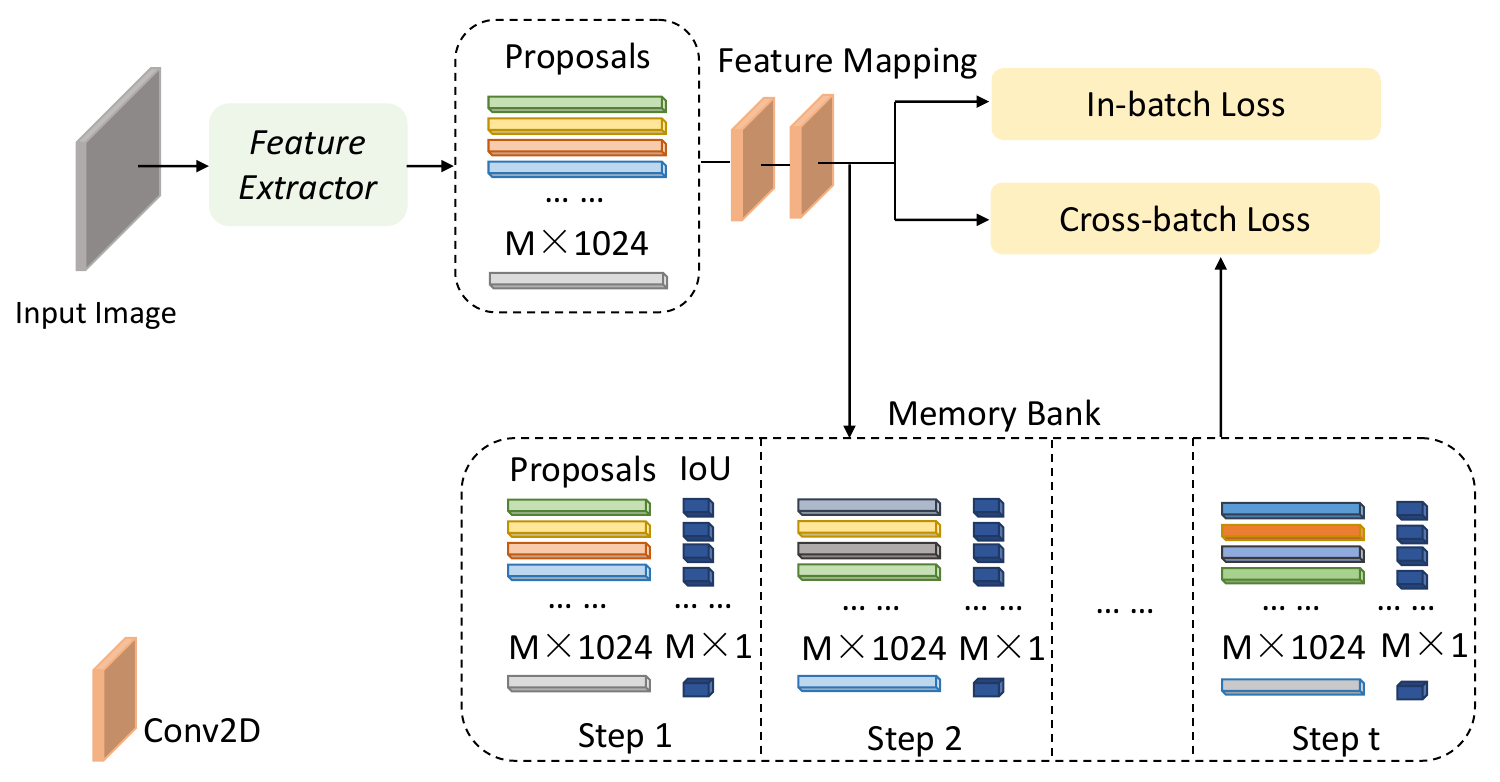}
  \caption{Memorable Contrastive Learning encoding (MCL) module. The memory bank stores for each proposal the feature embeddings and the IoU score between the proposal and the matched ground truth box.}
  \label{fig:MCL_encoder}
\end{figure}

The functional mechanism of our proposed MCL module is shown in Fig. \ref{fig:MCL_encoder}. The MCL module consists of a projection encoder and a memory bank. The projection encoder comprises 2 convolution layers with 1 ReLU activation layer between the convolution layers. It collects rotation-invariant features, maps proposal feature vectors to the embedding space, and stores them in the memory bank for subsequent contrastive learning.

\textbf{Object proposal memory bank.} In line with the MoCo framework \cite{He2020Moco}, we develop a memory bank tailored for storing object proposal information. This design marks a departure from conventional approaches that primarily store feature embeddings. Our object proposal memory bank is uniquely equipped to also hold the IoU scores, linking each object proposal with its corresponding ground truth bounding box, as well as the relevant ground truth labels. The implementation of this memory bank significantly expands the range of instances available for contrastive learning, see Fig. \ref{fig:MCL_encoder} for illustration. Consequently, this enhancement facilitates the model's ability to discern more distinct object features, particularly in a few-shot learning context. To accommodate this expanded data set, the capacity of the memory bank is deliberately configured to exceed the mini-batch size substantially.

\textbf{Contrastive loss.}\label{sec:ContrastiveLearning}
Drawing on the principles of supervised contrastive objectives in classification \cite{khosla2020supervised} and identification \cite{sun2014deep}, we formulate an instance-level contrastive loss based on batch-wise proposals and those in the memory bank. This loss is specifically designed to promote intra-class similarity while simultaneously enhancing inter-class separability, fostering a more robust and discriminative learning model. Specifically, given a batch of $M$ proposals, $\{x_i, c_i, y_i\}_{i= 1}^{M}$ where \(x_i\) is the orientation-invariant feature embeddings for $i$-th object proposal, \(c_i\) denotes the IoU between the proposal and its matched ground truth box, and \(y_i\) represents the ground truth label. All proposal features are normalized before calculating the contrastive loss.

Our MCL contrastive loss is formulated as:
\begin{equation}
\mathcal{L}_{MCL} = \frac{1}{M}\sum_{i = 1}^M \phi({c_i}) \cdot \mathcal{L}_i,
\end{equation}
where $\mathcal{L}_i$ denotes per-instance loss, and $\phi(c_i)$ is a weighting function that controls the consistency of proposals, calculated as,
\begin{equation}
    \phi(c_i) = \mathbb{I}(c_i>\theta) \cdot c_i
\end{equation}
where $\mathbb{I}$ denotes an identify function, and $\theta$ denotes the IoU threshold. By default, we set $\theta$ to 0.5 in our experiments. In this way, only the most centered object proposals will be included for model training.

For each proposal in the training batch, we calculate both in-batch and cross-batch contrastive losses, i.e., 
\begin{equation}
    \mathcal{L}_i = \mathcal{L}_{in} + \mathcal{L}_{cross}
\end{equation}

The in-batch contrastive loss is calculated as,
\begin{equation}
\mathcal{L}_{in} = \frac{-1}{M \times M} \sum_{i = 1}^{M} \sum_{j = 1}^M \log \frac{\exp(z_i \cdot z_j / \tau)}{\sum_{k=1}^N \mathbb{I}(k\neq i) (z_i \cdot z_k / \tau)},
\end{equation}
and the cross-batch contrastive loss is calculated as,
\begin{equation}
\mathcal{L}_{cross} = \frac{-1}{M \times N} \sum_{i = 1}^{M} \sum_{j = 1}^N w_{ij} \cdot \log \frac{\exp(z_i \cdot z_j / \tau)}{\sum_{k=1}^N \mathbb{I}(k\neq i) (z_i \cdot z_k / \tau)},
\end{equation}
where $\tau$ denotes a scale factor, and $N$ denotes the size of the memory bank. \({z_i}\) and \({z_j}\) denote the corresponding feature embeddings from the current training batch or those stored in the memory bank. $w_{ij}$ denotes the weight calculated as below.

In the original MoCo paper, a momentum update strategy is applied to the teacher encoder network and contrastive loss is calculated between feature embedding from the student and teacher networks. Similarly, in our MCL module, we employ a step-wise weighting approach to give different weights to proposals at different training time steps. Specifically, we put different weights for proposals in the memory banks, formulated as,
\begin{equation}
w_{ij} = \mathbb{I}(y_i = y_j) \cdot ({w_0} - \alpha  \cdot t_j),
\label{eq:membank_weight}
\end{equation}
where $t_j$ denotes the training step of the $j$-th proposal in the memory bank, counted backward from the current training step. $w_0$ is a hyperparameter that specifies the basis weight for the latest training iteration. \(\alpha \) represents the decay factor, and we set \(\alpha \) to ensure ${w_0} - \alpha \cdot t_j$ is always positive.

\begin{figure}
  \centering
  \includegraphics[width=0.48\textwidth]{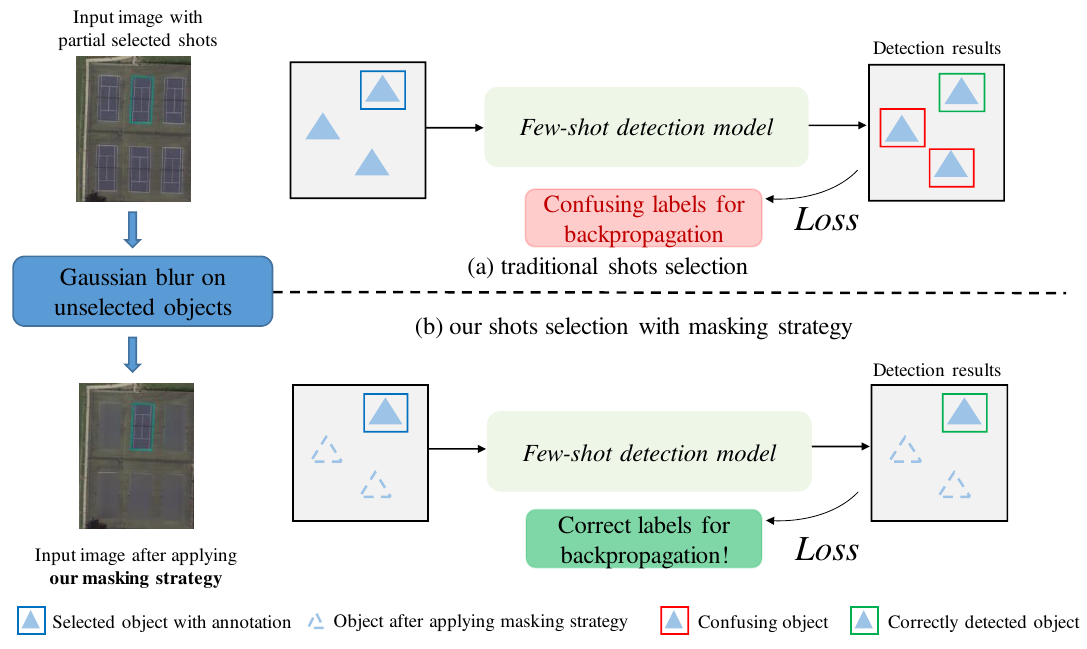}
  \caption{Illustration of the shot masking strategy employed in our work. Previous FSOD methods treat unselected objects as background and therefore cause confusion for model training. Our shot masking strategy masks out all unselected objects using a Gaussian blurring operation.} 
  \label{fig:dota_mask_strategy}
\end{figure}

\subsection{Masking Unchoosen Shots.}\label{sec_masking}

During the fine-tuning stage of few-shot object detection model training, only a limited number of shots are selected, which can lead to labeling errors. This issue arises when parts of objects from one image are chosen, while all other objects are designated as background, regardless of whether they belong to base or novel categories. This challenge is depicted in Fig. \ref{fig:dota_mask_strategy}(a), where instances with incorrect labels potentially confuse the few-shot detection model.

To address this problem, we implement a shot masking strategy in the data preparation phase. In scenarios where an image contains multiple object instances, but only a subset is selected for few-shot training, we apply a Gaussian blurring operation to mask out all unselected objects, irrespective of their category, see Fig. \ref{fig:dota_mask_strategy}(b). This technique ensures that these unselected instances are effectively excluded from the few-shot fine-tuning process, thereby reducing potential confusion and improving detection performance.

\subsection{Training Strategy}\label{sec_training}
In the first training phase, we train the base model for the detection of base categories. The overall loss consists of two parts, object classification loss (\({\mathcal{L}_{cls}}\)), and oriented bounding box regression loss (\({\mathcal{L}_{reg}}\)). 
After the base training phase, we modified the classification subnet to predict class labels in ${C_{base}} \cup {C_{novel}}$ during the fine-tuning phase.
During the fine-tuning stage, we froze the parameters of the backbone network and train other modules using the following loss functions:

\begin{equation}
\mathcal{L}_{total} = \mathcal{L}_{cls} + \mathcal{L}_{reg} + \lambda \mathcal{L}_{MCL},
\end{equation}
where \(\lambda \) is a hyperparameter used to balance the MCL loss with detection losses. Note that the MCL module is only used in the training stage and will not incur additional training costs in the inference stage.

\section{EXPERIMENTS AND RESULTS}

This section presents a comprehensive evaluation of our proposed FOMC on three remote sensing image datasets. First, we introduce the datasets used in our study and outline the implementation details, followed by optimal hyperparameter config. Then, we verify the superiority of  FOMC in terms of few-shot oriented object detection performance. Following this, we conduct extensive ablation studies and visualizations to demonstrate the effectiveness of each proposed component. Finally, we perform a comparative analysis of FOMC and the baseline model TFA to evaluate its performance on the FSOD task with HBBs. 

\begin{figure*}
  \centering
  \includegraphics[width=15.0cm]{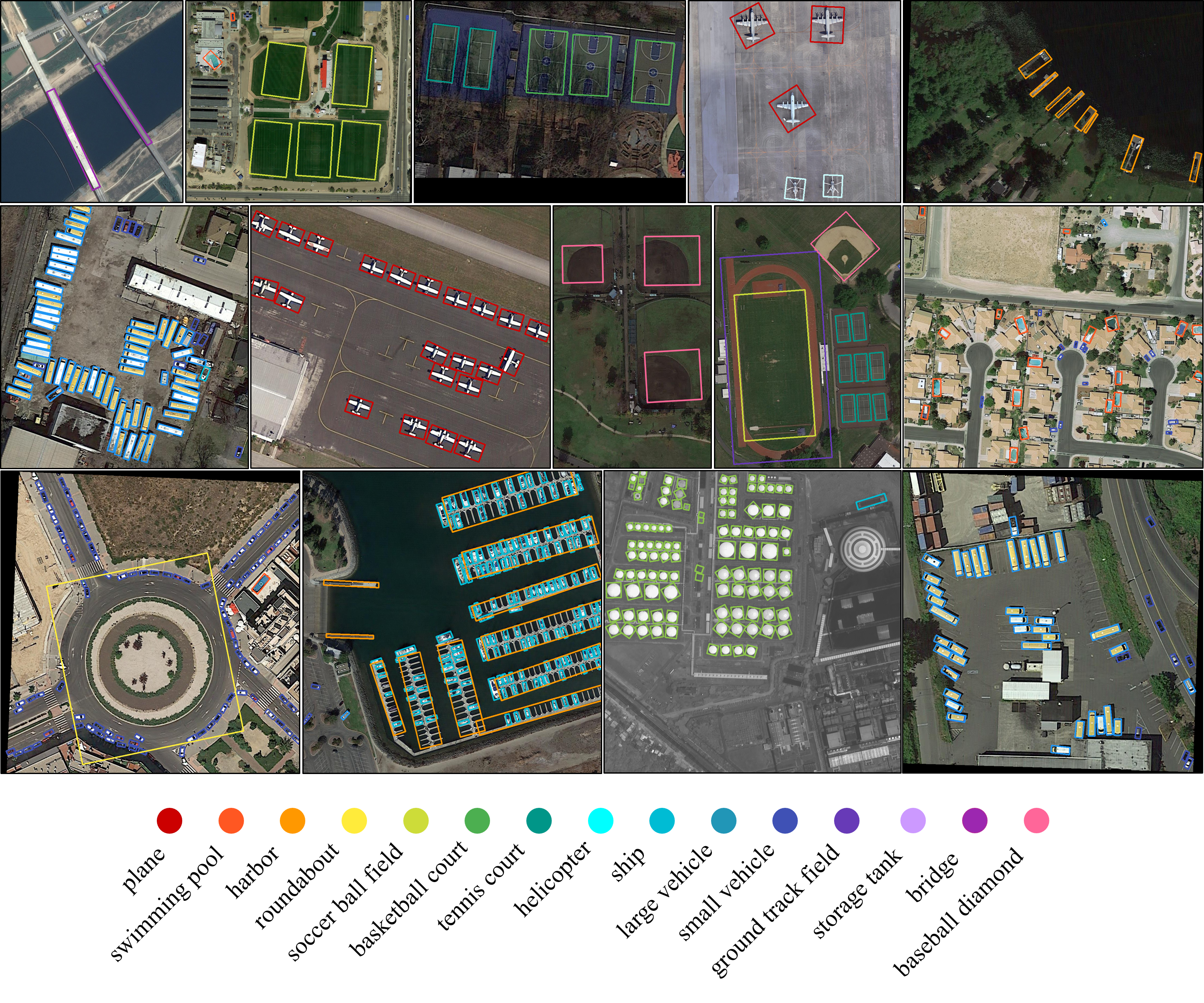}

  \caption{Detection results of DOTA. In the 20-shot setting, our proposed model FOMC achieves satisfactory performance in detecting oriented objects in complex backgrounds, crowded and arbitrary-oriented instances for both novel and base classes.} 
  \label{fig:dota few-shot result}

\end{figure*}

\begin{figure*}
  \centering
  \includegraphics[width=15.0cm]{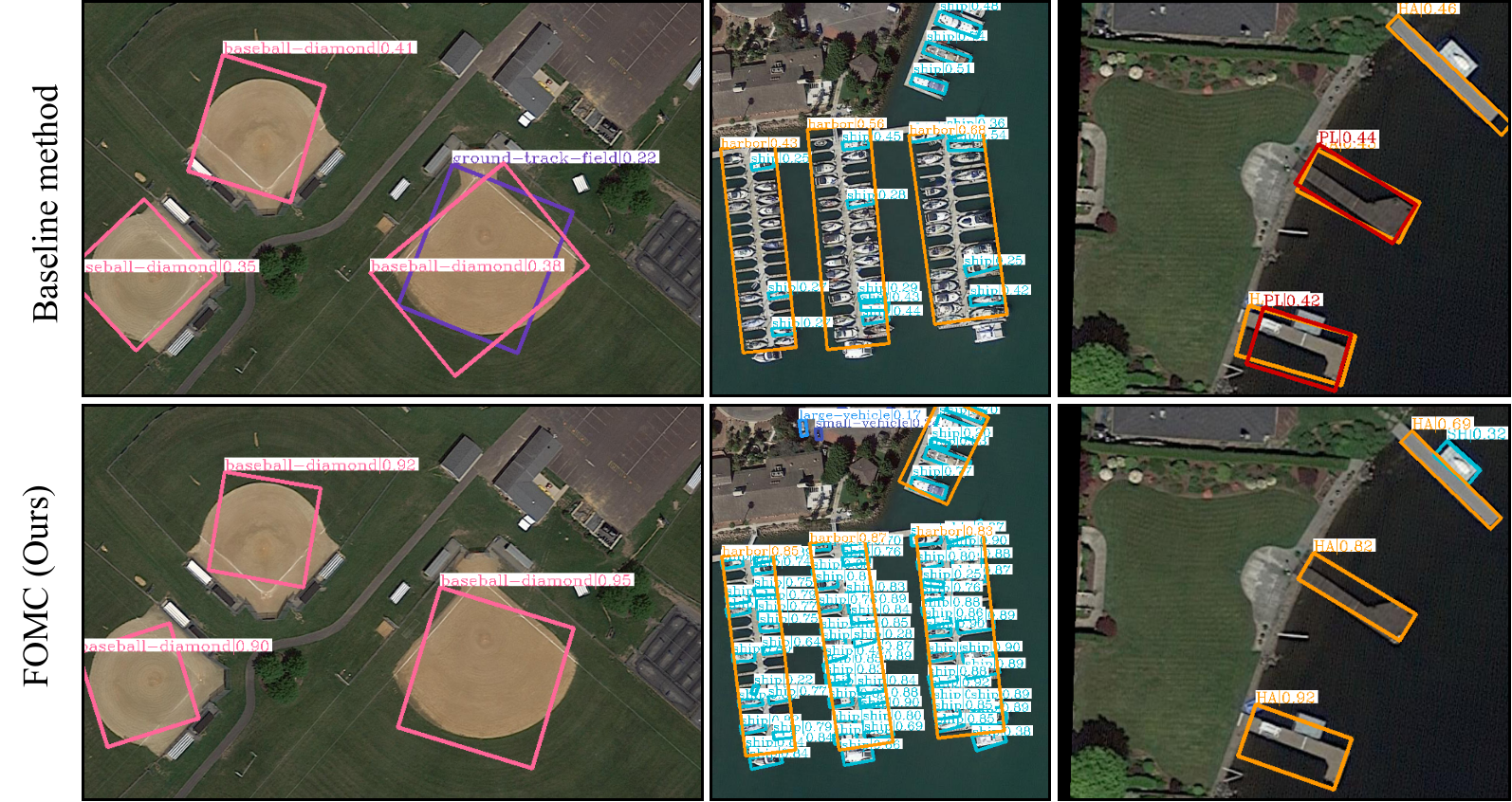}

  \caption{ Visualization of the few-shot detection results on the DOTA dataset with and without the MCL module. The baseline model struggles in class label prediction in the few-shot setting. Our MCL module explicitly captures intra-class similarity and inter-class differences, resulting in more accurate object detection in few-shot scenarios compared to the baseline method.} 
  \label{fig:dota_few_shot_compare}

\end{figure*}

\begin{figure*}
  \centering
  \includegraphics[width=18.0cm]{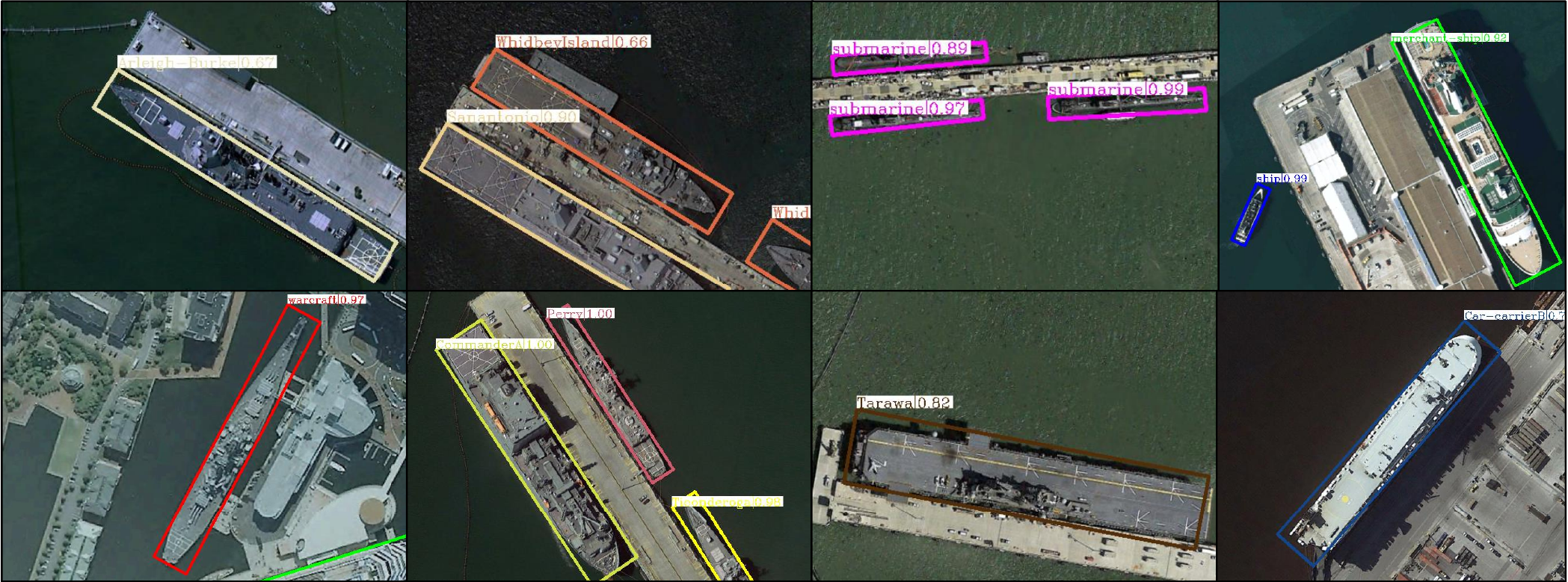}

  \caption{The visualization of few-shot oriented object ship detection results on the HRSC2016 dataset.} 
  \label{fig:hrsc2016_few-shot_result}

\end{figure*}

\subsection{Datasets}

\textbf{DOTA}~\cite{xia2018dota}. The DOTA benchmark dataset contains 2,806 aerial images with oriented object bounding box annotations. The size of these images ranges from 800 $\times$ 800 to 40000 $\times$ 40000 pixels. It consists of 188,282 instances of 15 categories: Plane (PL), Baseball-diamond (BD), Ground-track-filed (GTF), Small-vehicle (SV), Large-vehicle (LV), Ship (SH), Tennis-court (TC), Basketball-court (BC), Storage-tank (ST), Soccer-ball-field (SBF), Roundabout (RA), Harbor (HA), Swimming pool (SP), and Helicopter (HC). 1411 images of the dataset are randomly selected as the training set, 458 images as the validation set, and the remaining 937 images for testing. In this study, the original images are cropped into 1024 $\times$ 1024 sub-images with a stride of 824.

\textbf{HRSC2016}~\cite{cheng2016learning}. HRSC2016 is a challenging dataset designed for ship detection with OBB annotations. It has 1,061 images, containing a total of 2,976 instances of ships. The image sizes vary from 300 $\times$ 300 to 1500 $\times$ 900 pixels. It is a fine-grained dataset that includes three levels for ship detection tasks, namely L1, L2, and L3. We focus on the L3 task in this study, which involves 21 categories: Ship, Aircraft-carrier, Warcraft, Merchant-ship, Nimitz, Enterprise, Arleigh-Burke, WhidbeyIsland, Perry, Sanantonio, Ticonderoga, Container, Car-carrierA, Hovercraft, ContainerA, Submarine, Tarawa, Austen, CommanderA, Medical, and Car-carrierB. 

\textbf{NWPU VHR-10}~\cite{cheng2016learning}. It is a geospatial object detection dataset with HBBs covering ten categories: airplane, ship, storage tank, baseball diamond, tennis court, basketball court, ground track field, harbor, bridge, and vehicle. It consists of 800 optical images with resolutions ranging from 0.5 to 2m, of which 650 images are positive.

\subsection{Implementation Details }

We utilize the {\stanet} as the baseline network and adopt ResNet-50 \cite{he2016deep} with FPN as the backbone. For each level of pyramid features (P3 to P7), the predefined anchor sizes are typically chosen as \(\{ {32^2},{64^2},{128^2},{256^2},{512^2}\} \) pixels. The model is optimized by the standard stochastic gradient descent (SGD) optimizer with a momentum of 0.9 and a weight decay of 1e-4.
For the base training stage, the model is trained by 12 epochs on the DOTA dataset. The initial learning rate is set to 1e-2 and decreased by a factor of 0.1 at epoch 8 and 11. During the fine-tuning stage, the initial learning rate is set to 1e-3 and decreased by 0.1 at 8,000 iterations. A learning rate warm-up strategy is implemented for the first 200 iterations. The batch size is set to 8 during the base training stage and 12 during the fine-tuning stage. The hyperparameters of the MCL module are set as follows: \(\lambda  = 0.3\), \({w_0} = 0.95\), \(\alpha  = 0.01\), and the memory bank size is \(N = 8192\).

For the HRSC2016 dataset, we train the base model for 18 epochs. The initial learning rate is set to 1e-2 and decreases by a factor of 0.1 at epoch 12 and 16. During the fine-tuning stage, the initial learning rate is set to 1e-3 and decreases by 0.1 at 8,000 iterations. For the NWPU VHR-10 dataset, the base detector is trained for 24 epochs. The learning rate is set to 1e-2 and decreased by a factor of 0.1 at epoch 12 and 20. We also apply learning rate warmup for the first 100 iterations. 

In both base-training and fine-tuning stages, input images are augmented by randomly applying horizontal flipping and rotation. For the evaluation metrics, we utilize the $AP_{50}$ metric from VOC2007 benchmark~\cite{everingham2010pascal} to evaluate the performance of few-shot oriented object detection. Predictions with an IoU over 0.5 and correctly classified are considered true positives. All experiments are conducted using an Nvidia GeForce RTX 3090Ti GPU.

\subsection{Hyperparameter Selection}\label{sc_hyper}

\begin{table}
  \centering
  \setlength \tabcolsep{3pt}
    \caption{Comparison between the baseline model and our stronger baseline by replacing the last convolution layer with an FC layer.}
    \begin{tabular}{@{}cccc@{}}
        \toprule
        \multirow{2}{*}{FC} &  \multicolumn{3}{c}{mAP} \\ 
        \cline{2-4}
        & Base & Novel  & All \\
        
        \midrule
        \XSolid   & 0.70 & 0.84 & 0.73  \\
        \Checkmark & \textbf{0.72} & \textbf{0.87} & \textbf{0.76} \\
        \bottomrule
    \end{tabular}
 
  \label{table:s2anet_conv2fc_compare}
\end{table}

\noindent \textbf{Stronger baseline.}
In this study, we establish a stronger baseline model by simply replacing the last convolutional layer of the original classification branch with a fully-connected (FC) layer. Our motivation is to obtain instance-level feature vectors for subsequent contrastive learning. As reported in Table \ref{table:s2anet_conv2fc_compare}, the enhanced baseline outperforms the original \stanet model in the same few-shot setting, with an improvement of 3\% on mAP overall categories. In the following experiments, we utilize this stronger baseline model for performance comparison.

\begin{table}
  \centering
  
  \caption{Few-shot detection performance with different configures of memory back size (N) and loss weight ($\lambda$).}
  \begin{adjustbox}{width=0.7\linewidth}
    
    \begin{tabular}{cc|ccc|c}
      \toprule
      {N}  & $\lambda$  & PL & BD  & TC  & Avg. \\
      \midrule
      -       & -         & 0.33              & 0.39          & 0.53              & 0.41 \\
      \midrule
      0       & 0.05      & 0.37              & 0.30          & 0.56              & 0.41 \\
      0       & 0.3       & 0.40              & 0.36          & 0.66              & 0.47 \\
      0       & 0.5       & \textbf{0.44}     & 0.20          & 0.60              & 0.42 \\
      8192     & 0        & 0.37              & 0.25          & 0.61 & 0.41 \\
      8192     & 0.3       & 0.37              & \textbf{0.43} & \textbf{0.68} & \textbf{0.49} \\
      8192     & 0.5       & 0.40              & 0.38          & 0.60 & 0.46  \\
      4096     & 0.3       & 0.41              & 0.40          & 0.61 & 0.47  \\
      \bottomrule
    \end{tabular}
  \end{adjustbox}

  \label{table:dota_hyper_param_ablation}

\end{table}
\noindent \textbf{Effect of memory bank size.}
To study how our method performs with the hyperparameters, we train different settings for the memory bank size $N$ in the MCL module and evaluate the performance on the DOTA datasets in 20-shot settings. In our experiments, we try different memory bank sizes of 0, 4096, and 8192. Based on the ablative results reported in Table \ref{table:dota_hyper_param_ablation}, we observe that the detection accuracy improves as the memory bank size grows, with \(N = 8192\) leading to the best performance. Our contrastive learning method benefits from a larger memory bank size. We do not choose a further larger $N$ because this will add more computation cost during training.

\noindent \textbf{Effect of loss weight $\lambda$  in the MCL module.}
We further investigate how the detection performance changes using different loss weights in our MCL module. We conduct experiments with ($\lambda$) set to 0.05, 0.3, and 0.5. Table \ref{table:dota_hyper_param_ablation} shows that the performance increases from 0.05, plateaus at $\lambda=0.3$, and saturates when $\lambda$ is set to 0.5. Based on these results, we set $N$ to 8192 and $\lambda$ to 0.3 in the following experiments.

\subsection{Few-Shot Oriented Object Detection Results}

1) DOTA: Table \ref{table:dota_novel_compare_results} presents the performance comparison on the novel classes of the DOTA dataset. Our method outperforms the baseline \stanet model by  \(20\% \),  \(20\% \), and \(26\% \) in the 5-shot, 10-shot, and 20-shot settings, respectively. Moreover, as shown in Table \ref{table:s2anet_dota_base_results}, our method achieves an mAP that is \(15\% \) higher than that of {\stanet} on the base categories. These comparative studies demonstrate that our proposed model can effectively detect oriented objects and achieve satisfactory performance on novel classes without compromising the performance on the base classes. In addition to the above quantitative analysis, we also present the visualization of oriented object detection results in Fig \ref{fig:dota few-shot result}.

2) HRSC2016: The quantitative results on the HRSC2016 dataset are listed in the Table \ref{table:hrsc s2anet base training } and Table \ref{table:hrsc_novel_result}. For novel class object detection, our method achieves an mAP of \(47\% \), \(70\% \), and \(80\% \) under 3-shot, 5-shot, and 10-shot settings, respectively. Furthermore, our method gets satisfying performance on base categories with an mAP of \(77\% \). These results further validate that our method enhances novel class performance while also maintaining satisfactory detection performance on the base classes. The oriented ship detection results under the 10-shot setting are visualized in Fig. \ref{fig:hrsc2016_few-shot_result}.

\begin{figure}
  \centering
  \includegraphics[width=10.0cm]{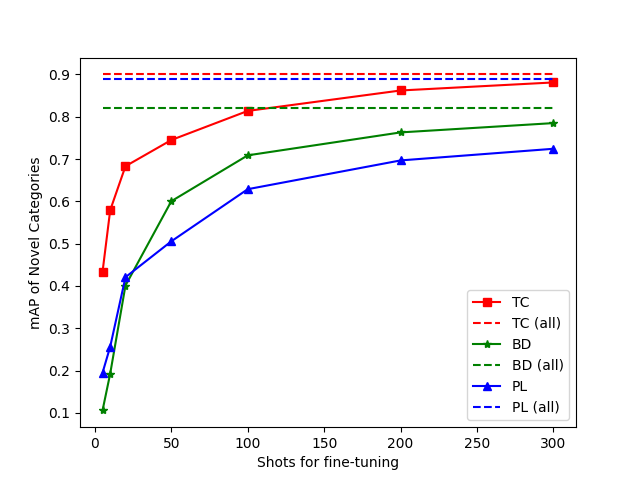}
  \caption{Few-shot detection performance on the novel categories of DOTA dataset with different numbers of shots. It is obvious that the mAP rapidly improves as the number of shots increases from 0 to 20, and then gradually saturates as the shots approach 200.}
  \label{fig:dota_novel_shot_results}
\end{figure}

\begin{figure}[h]
  \centering      \includegraphics[width=10cm]{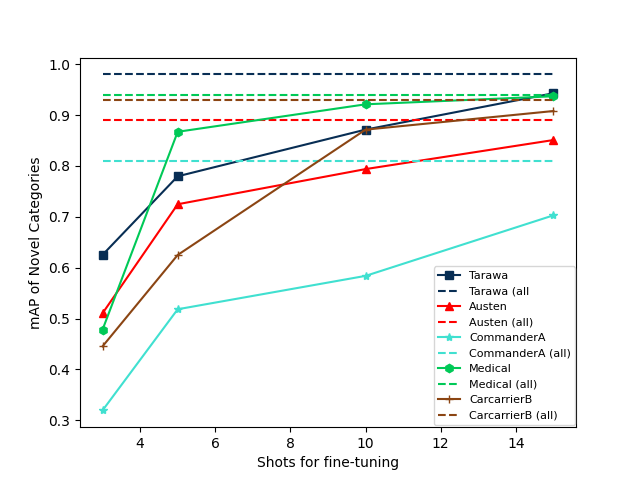}
  
  \caption{HRSC fine-tune results with different shots and reference precision when training with full HRSC2016 datasets}
  \label{fig:HRSC2016 results shots_ap}
\end{figure}

\begin{table}
  \centering
  \setlength\tabcolsep{3pt}
    \caption{Few-shot oriented object detection performance on novel categories of the DOTA dataset.}
    
    \begin{tabular}{@{}lcccccc@{}}
      \toprule
      \multirow{2}{*}{\textbf{~~~Method} } & \multicolumn{3}{c} {\stanet} & \multicolumn{3}{c}{FOMC}\\ 
      \cline{2-4} \cline{5-7}
      & 5-shot   & 10-shot    & 20-shot    & 5-shot        & 10-shot       & 20-shot       \\
      \midrule
      plane          & 0.13  & 0.22   & 0.33    & \textbf{0.20} & \textbf{0.26} & \textbf{0.37} \\
      baseball-diamond  & 0.13  & 0.17   & 0.39    & \textbf{0.11} & \textbf{0.20} & \textbf{0.43} \\
      tennis-court      & 0.34  & 0.41   & 0.53    & \textbf{0.43} & \textbf{0.58} & \textbf{0.68} \\
      \midrule
      Avg.              & 0.20  & 0.27   & 0.41    & \textbf{0.25} & \textbf{0.34} & \textbf{0.49} \\
      \bottomrule
    \end{tabular}

  \label{table:dota_novel_compare_results}

\end{table}

\begin{table}
  \centering
  \setlength\tabcolsep{3pt}
  \caption{Few-shot oriented object detection performance on base categories of the DOTA dataset.}
  \begin{tabular}{ @{}lccccccccccccccc@{} }
    \toprule
    \multirow{2}{*}{\textbf{~~~Method} } & \multicolumn{3}{c}{{\stanet}} & \multicolumn{3}{c}{FOMC}                                                                 \\
    \cline{2-4}                                     \cline{5-7}
    & 5-shot   & 10-shot     & 20-shot       & 5-shot        & 10-shot       & 20-shot       \\
    \midrule
    bridge                            & 0.29    & 0.26     & 0.27   & 0.44          & 0.46          & \textbf{0.49} \\
    ground-track-field                & 0.49    & 0.64     & 0.55   & 0.70          & 0.74          & \textbf{0.76} \\
    small-vehicle                     & 0.46    & 0.36     & 0.49   & 0.67          & \textbf{0.68} & \textbf{0.68} \\
    large-vehicle                     & 0.77    & 0.78     & 0.78   & 0.82   & 0.83 & \textbf{0.84} \\
    ship                              & 0.37    & 0.77     & 0.38   & 0.78          & 0.85 & \textbf{0.88} \\
    basketball-court                  & 0.62    & 0.59     & \textbf{0.66}   & 0.52 & 0.56          & 0.60          \\
    storage-tank                      & 0.83    & 0.75     & 0.83          & 0.87 & \textbf{0.88} & \textbf{0.88} \\
    soccer-ball-field                 & 0.59    & 0.66     & \textbf{0.69} & 0.64 & 0.64 & 0.67 \\
    roundabout                        & 0.56    & 0.49     & 0.58          & 0.69 & \textbf{0.71} & 0.70 \\
    harbor                            & 0.52    & 0.27     & 0.66          & 0.74 & 0.75 & \textbf{0.76} \\
    swimming-pool                     & 0.56    & 0.58     & 0.59          & 0.57 & 0.62 & \textbf{0.66} \\
    helicopter                        & 0.16    & 0.48     & 0.55          & 0.54 & 0.54 & \textbf{0.56}          \\
    \midrule
    Avg.                              & 0.52    & 0.55     & 0.61          & 0.66 & 0.69 & \textbf{0.70} \\
    \bottomrule
  \end{tabular}
  \label{table:s2anet_dota_base_results}
\end{table}

\begin{table}
  \centering
  \caption{Few-shot Oriented Object detection performance on novel classes of HRSC2016 dataset.}
  \begin{tabular}{lcccccc}
   \toprule
   \multirow{2}{*}{\textbf{~~~Method} } & \multicolumn{3}{c}{{\stanet}} & \multicolumn{3}{c}{FOMC} \\
   \cline{2-4}                                     \cline{5-7}
    & 3-shot & 5-shot & 10-shot & 3-shot & 5-shot & 10-shot \\ 
   \cline{1-7}
   Tarawa       & 0.36 & 0.69 & 0.87 & \textbf{0.63} & \textbf{0.78}  & 0.87 \\
   Austen       & 0.43 & 0.65 & 0.76 & \textbf{0.51} & \textbf{0.72}  & \textbf{0.79} \\
   CommanderA   & 0.21 & 0.50 & 0.55 & \textbf{0.32} & \textbf{0.52}  & \textbf{0.58} \\
   Medical      & 0.34 & 0.75 & 0.89 & \textbf{0.48} & \textbf{0.87}  & \textbf{0.92} \\
   Car-carrierB & 0.40 & 0.63 & 0.84 & \textbf{0.45} & 0.63  & \textbf{0.87} \\
   \midrule
   Avg.         & 0.35 & 0.65 & 0.78 & \textbf{0.48} & \textbf{0.70}  & \textbf{0.81} \\
   \bottomrule
  \end{tabular}
  
  \label{table:hrsc_novel_result}
\end{table}

\begin{table}
  \centering
  \caption{Few-shot Oriented Object detection performance on base classes of HRSC2016 dataset}
  \begin{tabular}{lc|lc}
   \toprule
    Category      & $AP_{50}$  & Category         & $AP_{50}$ \\ 
    \cline{1-4}
    Ship          & 0.65 & Aircraft-carrier & 0.84 \\
    Warcraft      & 0.27 & Merchant-ship    & 0.60 \\
    Nimitz        & 0.85 & Enterprise       & 0.40 \\
    Arleigh-Burke & 0.88 & WhidbeyIsland    & 0.79 \\
    Perry         & 0.90 & Sanantonio       & 0.96 \\
    Ticonderoga   & 0.90 & Container        & 0.77 \\
    Car-carrierA  & 0.75 & Hovercraft       & 0.66 \\
    ContainerA    & 0.84 & Submarine        & 0.91 \\
    \hline
    \multicolumn{2}{c}{Avg.} & \multicolumn{2}{c}{0.77} \\
   \bottomrule
  \end{tabular}
  
  \label{table:hrsc s2anet base training }

\end{table}

\begin{table}
  \centering
  \setlength \tabcolsep{3pt}
  
  \caption{Ablative performance of the MCL module and the masking strategy proposed in our method}
  
    \begin{tabular}{lcccccc}
      \toprule
      \multirow{2}{*}{\textbf{~~~Method} } & \multirow{2}{*}{mask} & \multirow{2}{*}{MCL} & \multicolumn{4}{c}{20-shot} \\ 
              \cline{4-7}
           &   &   & {PL}  & {BD} & {TC} & {Avg.}  \\
      \midrule
       Baseline & \XSolid    & \XSolid     & 0.25  & 0.26  & 0.57 & 0.36 \\
      \midrule   
       \multirow{3}{*}{~~~Ours} & \Checkmark & \XSolid  & 0.33  & 0.39  & 0.53 & 0.41 \\
        & \XSolid    & \Checkmark  & 0.32  & 0.34  & 0.61 & 0.43 \\
        & \Checkmark & \Checkmark  & \textbf{0.41}  & \textbf{0.43}  & \textbf{0.68} & \textbf{0.49} \\
      \bottomrule
    \end{tabular}

  \label{table:dota s2anet novel_category_shot results}

\end{table}

\begin{figure*}
  \centering
  \includegraphics[scale=0.5]{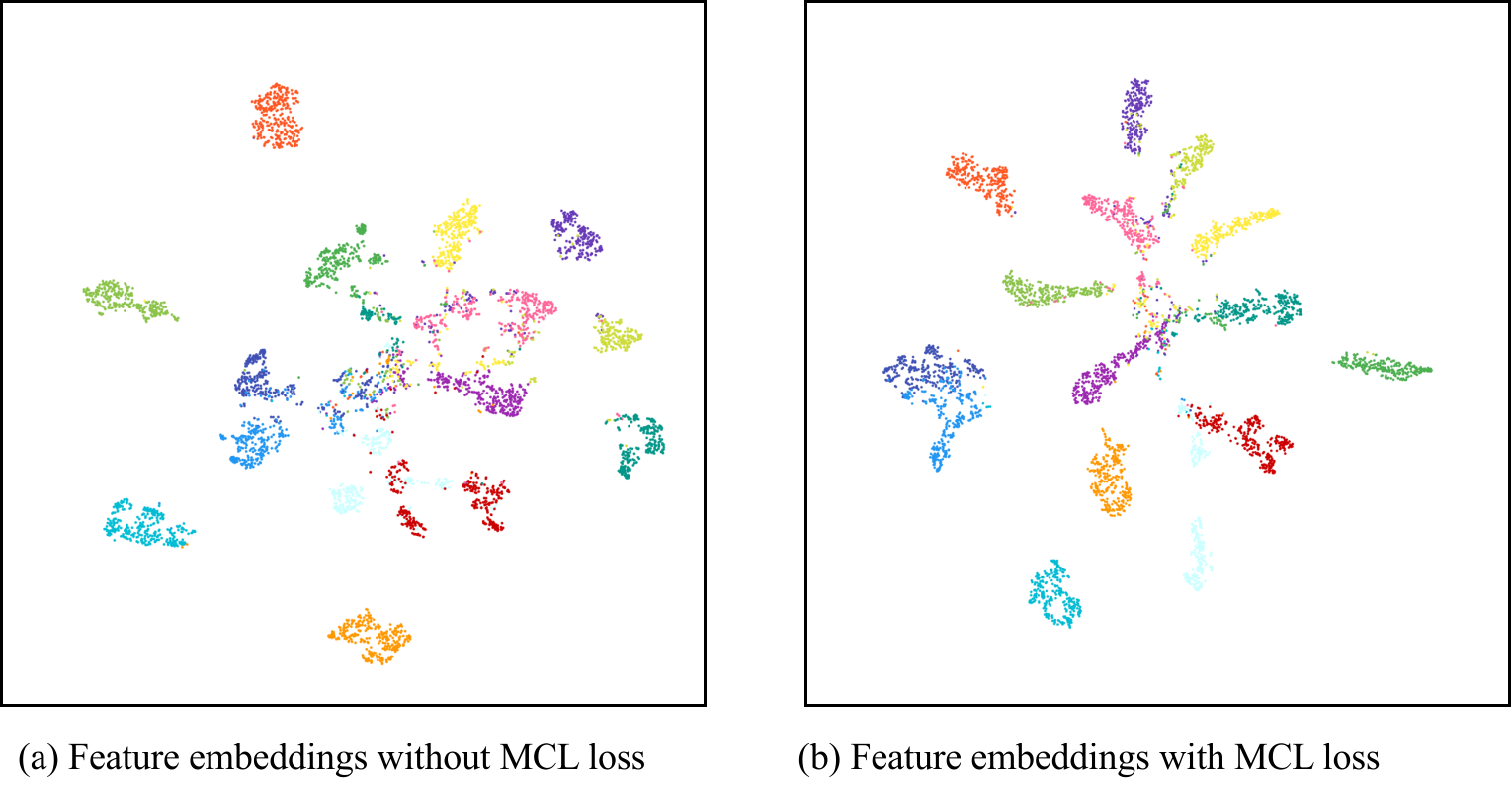}

  \caption{t-SNE visualization of proposal embeddings obtained with and without our proposed MCL loss in the latent space. The embedding space learned with the MCL loss exhibits a more compact intra-class feature distribution and larger inter-class margin.  } 
  \label{fig:dota_tsne_result}
\end{figure*} 

\begin{table*}
    \centering
    \caption{FEW-SHOT DETECTION PERFORMANCE ON THE NOVEL CLASSES OF THE NWPU VHR-10 DATASET.}
    \begin{tabular}{lccccccccc}
        \toprule
        \multirow{2}{*}{\textbf{~~~Method} } & \multicolumn{3}{c}{TFA\cite{wang2020frustratingly}} & \multicolumn{3}{c}{Li et al.\cite{li2023few}} & \multicolumn{3}{c}{FOMC}  \\
            \cline{2-10}
             & 3-shot & 5-shot & 10-shot & 3-shot & 5-shot & 10-shot  & 3-shot & 5-shot & 10-shot  \\ 
            \cline{1-10}
        airplane     & 0.12     & 0.51 & 0.60 & \textbf{0.41} & 0.53 & \textbf{0.77}  & 0.38  & \textbf{0.56}  & 0.74 \\
        baseball-diamond & 0.61 & 0.78 & 0.85 & 0.66 & 0.81 & \textbf{0.90}  & \textbf{0.73}  & \textbf{0.85}  & \textbf{0.90} \\
        tennis-court & 0.13     & 0.19 & 0.49 & \textbf{0.29} & \textbf{0.49} & \textbf{0.68}  & 0.17  & 0.46  & 0.59 \\
        \midrule
        Avg.         & 0.29     & 0.49 & 0.65 & \textbf{0.45} & 0.61 & \textbf{0.78}  & 0.43  & \textbf{0.62}  & 0.74 \\
        \bottomrule
    \end{tabular}
    
    \label{table:nwpu_few_shot_compare}
\end{table*}

\subsection{Ablation Study}
As shown in Table \ref{table:dota_novel_compare_results} and Table \ref{table:s2anet_dota_base_results}, our FOMC model exhibits enhanced capabilities in the detection of novel classes, while concurrently preserving the performance of base classes, as compared to the baseline 
\stanet model. We further compare our FOMC method with the stronger baseline (designed in Section \ref{sc_hyper}) to validate the effectiveness of our newly designed MCL module masking strategy. We conduct comparative experiments on the DOTA dataset and results are listed in Table \ref{table:dota s2anet novel_category_shot results}.

\noindent \textbf{Effect of MCL.} The MCL module achieves a \(19\% \) improvement on the novel classes compared with our stronger baseline model. Our MCL loss guides the model to establish contrastive-aware object feature embeddings. Learning discriminative embeddings with intra-class similarity and inter-class diversity eases the classification task for the few-shot detector. All ablation studies are done on the DOTA dataset in the 20-shot setting.  Fig. \ref{fig:dota_few_shot_compare} further illustrates the comparison results between our proposed method and the baseline \stanet model. Our memorable contrastive learning module enables FOMC to effectively capture the underlying data structure of the dataset, leading to a significantly improved and robust performance in the few-shot oriented object detection task.

 \noindent \textbf{Effect of shot masking strategy.}
In the fine-tuning stage, there are occasions when one image contains multiple foreground objects. In this case, only a part of the objects is provided with annotations, while other objects will be treated as background and hurt the learning process. We propose the shot masking strategy to mitigate the negative impact and present the ablative results in Table \ref{table:dota s2anet novel_category_shot results}. The shot masking strategy improves the accuracy from \(36\% \) to \(41\% \), demonstrating its effectiveness in boosting few-shot detection performance. 

\noindent \textbf{Robustness to random shot selection.} 
In order to demonstrate the robustness of our few-shot detection model in the face of shot selection randomness, we repeated the experiments ten times, each time selecting shots from novel categories in a different random manner. These experiments were carried out using the DOTA dataset, employing a 20-shot configuration. Our FOMC model consistently achieved an average mean Average Precision (mAP) of $47.3\%\pm2.6\%$ for the novel categories and $66.7\%\pm1.5\%$ for the base categories. These results serve as a compelling indication of the robustness of our FOMC model to the variability in shot selection.

\noindent \textbf{Number of shots.}
We study the performance of our few-shot oriented object detection model FOMC with different numbers of shots on novel categories. As shown in Fig. \ref{fig:dota_novel_shot_results}, the performance rapidly improves as the number of shots increases from 0 to 20, and then gradually saturates as the shots approach 200. In the case of the tennis court class, our proposed FOMC model achieves an mAP of about \(80\% \) with only 100 shots, and nearly \(90\% \) with 200 shots, which is comparable to the baseline model training with the full set of novel examples. However, for the airplane classes, the oriented object detection task is more challenging in a limited-shot setting. This is likely due to the fact that airplanes have significant intra-class variance in terms of size, orientation, and appearance. Even though, our FOMC model can effectively detect oriented aerial objects with limited examples and achieves comparable performance to the baseline model given abundant data. This conclusion is further supported by our experiments on the HRSC2016 dataset. Figure \ref{fig:HRSC2016 results shots_ap} illustrates that our model learns from only 10 shots to achieve a similar accuracy to the baseline trained with the whole training samples for the medical ship class. For the more challenging oriented ship detection task, our FOMC model can learn robust and discriminative features to recognize ships from fine-grained categories with only a few samples.

\subsection{t-SNE visualization}
We select the data-rich DOTA dataset for visualization experiments. As shown in Fig. \ref{fig:dota_tsne_result}, the embedding space obtained by the baseline model is unable to distinguish proposal embeddings from different categories, due to the overlapped and entangled distributions of data points. With the help of our proposed MCL module, the feature embeddings from the same categories are more tightly clustered, and the margin between different categories is enlarged. As a result, the model learns to capture the underlying semantics and achieves better performance in the few-shot oriented object detection task.

\subsection{Few-shot Object Detection Results}
We conduct additional experiments on the NWPU VHR-10 dataset to show the effectiveness of our method for conventional few-shot object detection with horizontal bounding boxes. We compare our model FOMC with the prevalent fine-tuning-based method TFA \cite{wang2020frustratingly} and the state-of-the-art method \cite{li2023few}. It is worth mentioning that both TFA and \cite{li2023few} are based on the two-stage object detection framework Faster R-CNN, which gives them an edge in terms of accuracy but compromises computational speed. As shown in Table \ref{table:nwpu_few_shot_compare}, our proposed FOMC outperforms the TFA method, improving the performance by \(10\% \) in all settings. In the 5-shot setting, FOMC achieves an mAP of  \(62\% \) for novel classes. In 3-shot and 10-shot settings, FOMC is comparable to the state-of-the-art performance, demonstrating its strong capability in conventional few-shot detection with HBBs.

\section{Conclusion}

In this study, we propose the FOMC model aiming to address the more challenging oriented object detection task in few-shot settings. Furthermore, we design the MCL module, a contrastive learning-based approach with a memorable queue to learn more discriminative features, further improving the detection performance for overall classes. To mitigate the negative impact of incompletely labeled objects during the fine-tuning stage, we further propose the masking strategy. We conduct extensive experiments on two public benchmark datasets DOTA and HRSC 2016 to validate the effectiveness of our proposed methods. To our best knowledge, we are the first to solve few-shot oriented object detection and we will further explore how to enhance its performance for more complex situations.

\appendices

\section*{Acknowledgment}

The project is supported by the Special Fund of Hubei Luojia Laboratory (No.220100013)

\nolinenumbers
\bibliographystyle{unsrt} 


\newpage

\setcounter{equation}{0}
\renewcommand\theequation{A.\arabic{equation}}
\setcounter{figure}{0}
\renewcommand\thefigure{A.\arabic{figure}}
\setcounter{table}{0}
\renewcommand\thetable{A.\arabic{table}}



\end{document}